\documentclass[sigconf]{acmart}

\usepackage[utf8]{inputenc} %
\usepackage[T1]{fontenc}    %
\usepackage{url}            %
\usepackage{booktabs}       %
\usepackage{amsfonts}       %
\usepackage{nicefrac}       %
\usepackage{microtype}      %

\usepackage{color,xcolor}
\usepackage{graphicx}
\usepackage{framed}
\usepackage{array}
\usepackage{booktabs}
\usepackage{colortbl}
\usepackage{float,wrapfig}
\usepackage{hhline}
\usepackage{multirow}
\usepackage{subcaption} %
\usepackage[labelfont={bf},labelsep={period},font={small}]{caption}

\usepackage{tabu}

\let\llncssubparagraph\subparagraph
\let\subparagraph\paragraph
\let\subparagraph\llncssubparagraph

\usepackage{amsmath,amsfonts}
\usepackage{bm}
\usepackage{nicefrac}
\usepackage{mathtools}

\usepackage{changepage}
\usepackage{extramarks}
\usepackage{fancyhdr}
\usepackage{setspace}
\usepackage{soul}
\usepackage{xspace}

\usepackage{url}

\usepackage{algorithm}
\usepackage{algpseudocode}
\usepackage{enumerate}

\usepackage{pbox}
\usepackage{tablefootnote}

\definecolor{mydarkgreen}{rgb}{0.02,0.6,0.02}

\usepackage{ dsfont }

\usepackage{pifont}
\usepackage{nccmath}

\usepackage{listings}
\usepackage{xcolor}

\definecolor{codegreen}{rgb}{0,0.6,0}
\definecolor{codegray}{rgb}{0.5,0.5,0.5}
\definecolor{codepurple}{rgb}{0.58,0,0.82}
\definecolor{backcolour}{rgb}{0.95,0.95,0.92}

\lstdefinestyle{mystyle}{
    backgroundcolor=\color{backcolour},   
    commentstyle=\color{codegreen},
    keywordstyle=\color{magenta},
    numberstyle=\tiny\color{codegray},
    stringstyle=\color{codepurple},
    basicstyle=\ttfamily\footnotesize,
    breakatwhitespace=false,         
    breaklines=true,                 
    captionpos=b,                    
    keepspaces=true,                 
    numbers=left,                    
    numbersep=5pt,                  
    showspaces=false,                
    showstringspaces=false,
    showtabs=false,                  
    tabsize=2
}
\usepackage{hyperref}

\definecolor{mydarkblue}{rgb}{0,0.08,1}

\newcommand{\myparagraph}[1]{\vspace{0pt}\paragraph{\textbf{#1}}}
\DeclareCaptionType{InfoBox}

\newcommand{\ignorethis}[1]{}

\makeatletter
\DeclareRobustCommand\onedot{\futurelet\@let@token\@onedot}
\def\@onedot{\ifx\@let@token.\else.\null\fi\xspace}

\def\eg{\emph{e.g}\onedot}

\def\etc{\emph{etc}\onedot} 
\def\wrt{w.r.t\onedot} 

\makeatother

\makeatletter

\newcommand\footnoteref[1]{\protected@xdef\@thefnmark{\ref{#1}}\@footnotemark}
\makeatother

\definecolor{fig5gray}{rgb}{0.80, 0.85, 0.89}
\definecolor{fig5red}{rgb}{1.00, 0.59, 0.55}
\definecolor{fig5yellow}{rgb}{1.00, 0.68, 0.00}

\definecolor{mydarkblue}{rgb}{0,0.08,1}
\definecolor{mydarkred}{rgb}{0.8,0.02,0.02}
\definecolor{mydarkorange}{rgb}{0.40,0.2,0.02}
\definecolor{mypurple}{RGB}{111,0,255}
\definecolor{myred}{rgb}{1.0,0.0,0.0}
\definecolor{mygold}{rgb}{0.75,0.6,0.12}
\definecolor{mydarkgray}{rgb}{0.66, 0.66, 0.66}
\definecolor{mygray}{gray}{0.9}

\def\engine{PockEngine\xspace}
\def\engineshort{PockEngine\xspace}

\newcommand{\cmark}{\ding{51}}%
\newcommand{\xmark}{\ding{55}}%

\usepackage{color}
\usepackage{soul}

\definecolor{lightgrey}{rgb}{0.925, 0.925, 0.925}
\sethlcolor{lightgrey}

\makeatletter
\def\SOUL@hlpreamble{%
    \setul{}{2.5ex}%
    \let\SOUL@stcolor\SOUL@hlcolor
    \dimen@\SOUL@ulthickness
    \dimen@i=-.75ex %
    \advance\dimen@i-.5\dimen@
    \edef\SOUL@uldepth{\the\dimen@i}%
    \let\SOUL@ulcolor\SOUL@stcolor
    \SOUL@ulpreamble
}
\makeatother

\AtBeginDocument{%
  }

\setcopyright{acmcopyright}
\copyrightyear{2023}
\acmYear{2023}
\setcopyright{rightsretained}
\acmConference[MICRO '23]{56th Annual IEEE/ACM International Symposium on Microarchitecture}{October 28-November 1, 2023}{Toronto, ON, Canada}
\acmBooktitle{56th Annual IEEE/ACM International Symposium on Microarchitecture (MICRO '23), October 28-November 1, 2023, Toronto, ON, Canada}
\acmDOI{10.1145/3613424.3614307}
\acmISBN{979-8-4007-0329-4/23/10}

\begin{document}

\title[PockEngine]{PockEngine: Sparse and Efficient Fine-tuning in a Pocket}

%\author{Ligeng Zhu$^1$, Lanxiang Hu$^2$, Ji Lin$^1$, Wei-Chen Wang$^1$, Wei-Ming Chen$^1$, Chuang Gan$^3$,  Song Han$^{1, 4}$}
%\affiliation{
%    \institution{MIT$^1$, UCSD$^2$, MIT-IBM Watson AI Lab$^3$, NVIDIA$^4$}
%    \country{USA}
%}
%\email{{ligeng, jilin, wweichen, wmchen, chuangg, songhan}@mit.edu, hlxde2@gmail.com}

\author{Ligeng Zhu}
\affiliation{
    \institution{MIT}
    \city{Cambridge}
    \state{MA}
    \country{USA}
}
\email{ligeng@mit.edu}

\author{Lanxiang Hu}
\affiliation{
    \institution{UCSD}
    \city{San Diego}
    \state{CA}
    \country{USA}
\city{}
}
\email{hlxde2@gmail.com}

\author{Ji Lin}
\affiliation{
    \institution{MIT}
    \city{Cambridge}
    \state{MA}
    \country{USA}
\city{}
}
\email{jilin@mit.edu}

\author{Wei-Chen Wang}
\affiliation{
    \institution{MIT}
    \city{Cambridge}
    \state{MA}
    \country{USA}
\city{}
}
\email{wweichen@mit.edu}

\author{Wei-Ming Chen}
\affiliation{
    \institution{MIT}
    \city{Cambridge}
    \state{MA}
    \country{USA}
\city{}
}
\email{wmchen@mit.edu}

\author{Chuang Gan}
\affiliation{
    \institution{MIT-IBM Watson AI Lab}
    \city{Cambridge}
    \state{MA}
    \country{USA}
}
\email{chuangg@mit.edu}

\author{Song Han}
\affiliation{%
 \institution{MIT, NVIDIA}
  \city{Cambridge}
  \state{MA}
  \country{USA}
}
\email{songhan@mit.edu}

\renewcommand{\shortauthors}{Zhu et al.}

\begin{abstract}
    On-device learning and efficient fine-tuning enable continuous and privacy-preserving customization (e.g., locally fine-tuning large language models on personalized data).
    However, existing training frameworks are designed for cloud servers with powerful accelerators (\eg, GPUs, TPUs) and lack the optimizations for learning on the edge, which faces challenges of resource limitations and edge hardware diversity.
    We introduce \textbf{PockEngine}: a tiny, sparse and efficient engine to enable fine-tuning on various edge devices. 
    PockEngine supports \textbf{sparse backpropagation}: it prunes the backward graph and sparsely updates the model with measured memory saving and latency reduction while maintaining the model quality.  
    Secondly, PockEngine is \textbf{compilation first}: the entire training graph (including forward, backward and optimization steps) is derived at compile-time, which reduces the runtime overhead and brings opportunities for graph transformations.
    PockEngine also integrates a rich set of \textbf{training graph optimizations}, thus can further accelerate the training cost, including operator reordering and backend switching.
    \engineshort supports \textbf{diverse applications, frontends and hardware backends}: it flexibly compiles and tunes models defined in PyTorch/TensorFlow/Jax and deploys binaries to mobile CPU/GPU/DSPs. 
    We evaluated \engineshort on both vision models and large language models.
    \engineshort achieves up to 15 $\times$ speedup over off-the-shelf TensorFlow (Raspberry Pi), 5.6 $\times$ memory saving back-propagation (Jetson AGX Orin).
    Remarkably, PockEngine enables fine-tuning LLaMav2-7B on NVIDIA Jetson AGX Orin at 550 tokens/s, 7.9$\times$ faster than the PyTorch. 
\end{abstract}




\sloppy
\maketitle

\begin{figure}
    \centering
    \includegraphics[width=1\linewidth]{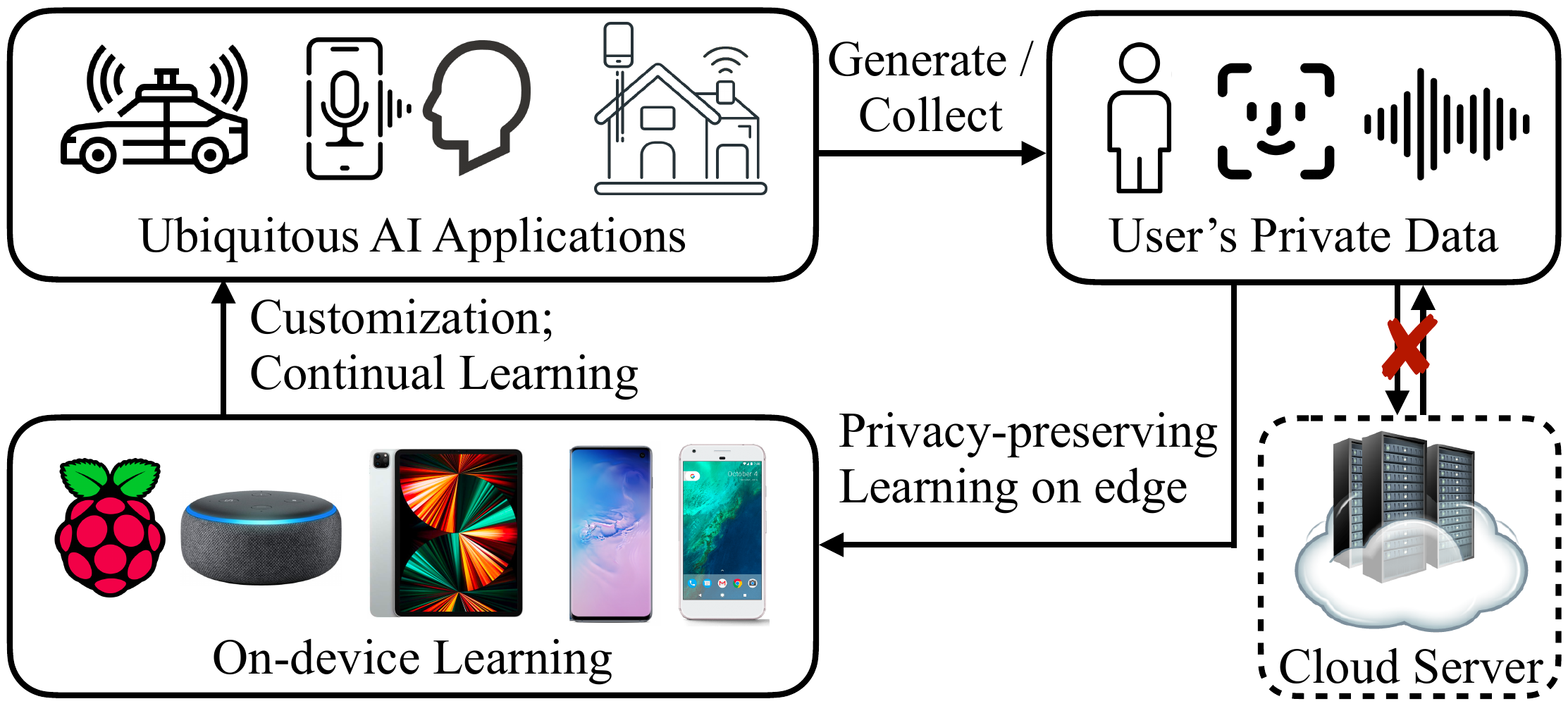}
    \caption{On-device learning and local fine-tuning enable customization, protect privacy, and form a virtuous cycle between user and devices. 
    }
    \label{fig:teaser}
\end{figure}

\section{Introduction}

Edge devices are ubiquitous and produce an increasing amount of data in our daily lives. The need for intelligent, personalized, and private AI is rapidly growing, as a single model fails to fit different users' needs.
However, while deep learning inferences are widely performed on edge devices, the training of deep neural networks is typically run on cloud GPU servers.
Cloud-based training requires users to upload their personal data to the cloud, which not only incurs additional data transfer costs, but also brings privacy risks over sensitive data (\eg, healthcare data, keyboard input history, GPS location, \etc).

On-device training is a promising solution for model customization without sacrificing privacy (Figure~\ref{fig:teaser}). It allows a pre-trained model to continuously adapt to sensor data without sending it to the cloud. For example, the smart keyboard model can update itself to better predict the next word from users' typing history; the email assistant can learn from users' previous drafts and train personalized language models; vision models can automatically adapt to environments with domain shifts~\cite{sun2020test}). 
The near-sensor training paradigm also brings important benefits for energy and connectivity: it saves energy from data transmission (which is much more expensive than computation~\cite{levis2004trickle}); it also helps with applications like ocean sensing~\cite{jang2019underwater} and smart agriculture~\cite{vasisht2017farmbeats} that do not have physical access to the Internet.

Despite all the benefits, on-device training is difficult due to the following challenges:

\textbf{(1) Resource Limitations.} 
The capacity of edge devices is orders of magnitude smaller than cloud servers. People have been trying hard to squeeze deep learning models just for edge \emph{inference}, while model \emph{training} and \emph{fine-tuning} are more power-, computation-, and memory-expensive. 
We need extra memory to store all intermediate feature maps for backpropagation, and extra computation for the backward pass (roughly $3\times$ compared to inference).
Sometimes the training needs a larger batch size to ensure a stable convergence, making the process even more costly.
For MobilenetV2~\cite{sandler2018mobilenetv2} the training memory is 14$\times$ and 7.3 $\times$ larger than inference (batch size 8) 
and for BERT~\cite{devlin2018bert} the peak memory usage is 7.3 $\times$ larger compared to inference. 
Furthermore, the optimizers also require extra memory (2x for Momentum and 3x for Adam~\cite{kingma2014adam}). 
With the current training framework, 
the training costs could soon exceed the resource limits of edge hardware. 

\textbf{(2) Hardware Diversity }
While the accelerators on cloud servers are dominated by GPUs, the hardware of edge platforms has a wide range of options on the market. The processor ranges from ARM microcontrollers to powerful Apple M1 chips, and the accelerator varies between Qualcomm Adreno GPUs, Hexagon DSPs, and edge TPUs. Each hardware comes with a different inference library. \engineshort can directly use these inference libraries for training by compiling the training graph into standard ONNX format.  
On the other hand, popular deep learning training frameworks like TensorFlow~\cite{tensorflow2015}, PyTorch~\cite{pytorch2019} and Jax~\cite{jax2018github} are developed for high-end cloud GPUs/TPUs.
The performance is poor when directly applied to edge platforms\footnote{The frameworks themselves cannot even be installed due to the tight resource constraints of low-end hardware like microcontrollers~\cite{lin2022ondevice}.}.

\begin{figure*}[htb]
    \centering
    \includegraphics[width=1\linewidth]{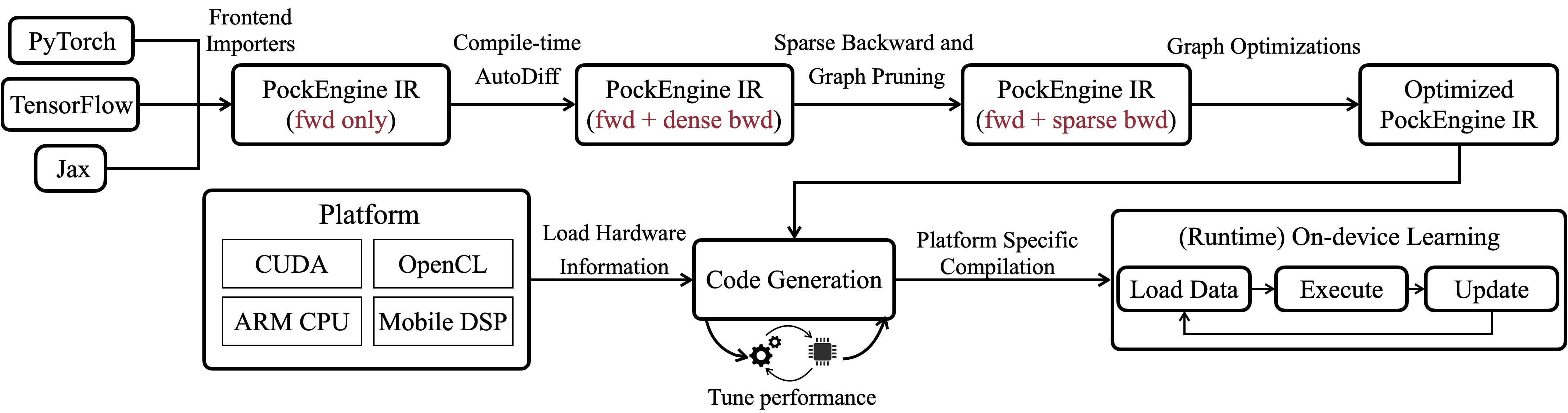}
    \caption{The workflow of \engine. 
    \engineshort performs the auto-diff at compile-time, prunes the computation graph to support sparse backpropagation, and enables previously inference-only hardware platforms to perform backpropagation. \engineshort enables efficient fine-tuning on resource-constrained devices like NVIDIA Jetson and mobile devices.
    }
    \label{fig:overview}
\end{figure*}

To address the above challenges, we introduce \textbf{\engineshort}, a tiny and efficient training engine designed for on-device training. We highlight the following properties:

\begin{itemize}
    \item \engineshort provides system-level support for both \textbf{dense and sparse backpropagation}. Apart from updating the whole model, \engineshort supports flexible \emph{sparse} update schemes by computing the gradients for only part of the weights, which proves to be a more efficient option for fine-tuning/transfer learning without harming the accuracy~\cite{cai2020tinytl, mudrakarta2018k, frankle2020training, lin2022ondevice, houlsby2019parameter, hu2021lora, li2021prefix}. Existing training frameworks can only \emph{simulate} the sparse backpropagation by computing the backward and mask out gradients, but cannot \emph{realize} measured speed up and memory savings. \engineshort supports sparse backpropagation via graph pruning and dead code elimination with the compilation nature, leading to smaller computation and memory usage.
    \item \engineshort is a \textbf{compilation-based} efficient training engine and enables many \textbf{inference-only framework to perform training}. Our compilation workflow helps to connect diverse model architectures and frontend options (\eg, vision/NLP models, PyTorch/TensorFlow/ONNX definitions) with various backend libraries (\eg, SNPE for Qualcomm, Metal for Apple Silicon, TVM), exposing a unified intermediate representation (IR). By sharing the same set of operators for both forward and backward operations, we not only enable inference frameworks to train neural networks, but also allow for various graph optimizations to improve efficiency (see Figure~\ref{fig:overview}).
    \item \engineshort implements a rich set of \textbf{graph optimizations} to improve the efficiency on edge devices, including operator fusion, operator reordering, layout transforms, and backend switching that are conventionally used for \emph{inference only}. We find that the training graphs actually have more optimization opportunities due to their complexity. By sharing the same operator set with inference graphs, \engineshort can well utilize the optimization techniques from inference engines (\eg, \engineshort utilizes previously inference-only winograd convolution to accelerate training).  
\end{itemize}

We extensively evaluated \engineshort on six edge platforms and six deep learning tasks from vision to NLP. PockEngine achieves up to 11$\times$ speedup over TensorFlow for the same training workload.  With sparse backpropagation, we can further improve the acceleration up to 21$\times$ without losing transfer learning accuracy on tiny microcontrollers. We hope our work can contribute to the thriving of on-device training by providing a \emph{general-purpose, high-efficiency, user-friendly} training framework for edge devices.

\begin{table*}[h]
\centering
\small
\caption{
    Comparison between existing deep learning frameworks. ``-'' denotes the feature is not fully supported for training. 
}
\label{tab:framework_comparison}
\begin{tabular}{c|c|c|c|c|c|c}
\toprule
& \begin{tabular}[c]{@{}c@{}}Support\\Training\end{tabular} 
& \begin{tabular}[c]{@{}c@{}}Support \\ Sparse-BP\end{tabular} 
& \begin{tabular}[c]{@{}c@{}}Run without\\ Host Language\end{tabular} 
& \begin{tabular}[c]{@{}c@{}}Kernel Optimized \\ for Edge\end{tabular} 
& \begin{tabular}[c]{@{}c@{}}Compile-Time \\ AutoDiff\end{tabular} 
& \begin{tabular}[c]{@{}c@{}}Graph \\ Optimizations\end{tabular} \\ \midrule
PyTorch~\cite{pytorch2019}              & \cmark & \xmark & \xmark & \xmark & \xmark & \xmark \\ \midrule
TensorFlow~\cite{tensorflow2015}           & \cmark & \xmark & \xmark & \xmark & \xmark & - \\ \midrule
Jax~\cite{jax2018github}           & \cmark & \xmark & \xmark & \xmark & \xmark & \xmark \\ \midrule
TVM~\cite{chen2018tvm}                  & \xmark & \xmark & \cmark & \cmark & - & \cmark \\ \midrule
MNN~\cite{jiang2020mnn}                  & \cmark & \xmark & \cmark & \cmark & \xmark & \xmark \\ \midrule
\engine (ours) & \cmark & \cmark & \cmark & \cmark & \cmark & \cmark \\ \bottomrule
\end{tabular}
\end{table*}

\section{Related Work}

\subsection{Cloud Deep Learning Systems}
The success of deep learning is built on top of popular training frameworks such as  PyTorch~\cite{pytorch2019}, TensorFlow~\cite{abadi2016tensorflow}, MXNet~\cite{chen2015mxnet}, JAX~\cite{jax2018github}, \etc. 
These systems are designed for development flexibility and depend on a host language (\eg, Python) to execute. This brings significant memory overhead (>300MB) and makes the runtime especially slow on low-frequency CPU (e.g., ARM Cortex). Moreover, the operator kernels are optimized for high-end GPU devices and lack performance tuning for edge devices and some overheads such as extra gradient buffers for the optimizer step are not considered a bottleneck for powerful server hardware.
\engineshort is a compilation-based framework thus the runtime does not rely on host languages as compared in Table~\ref{tab:framework_comparison}. This moves most workloads from runtime to compile-time to minimize the runtime overhead and enables later optimizations to improve training throughput.

\subsection{Edge Deep Learning Systems}
When deploying models on tiny edge devices, inference libraries like TVM~\cite{chen2018tvm}, TF-Lite, NCNN~\cite{ncnn}, TensorRT~\cite{tensorRT}, and OpenVINO~\cite{vaswani2017attention} 
deliver optimized kernels for mobile platforms and provide a lightweight runtime without host language. However, they focus mostly on inference and do not support on-device training. MNN~\cite{jiang2020mnn} has preliminary support for CNNs but the flexibility is rather limited and it does not optimize training memory usage.
POET~\cite{patil2022poet} applies rematerialization and paging to deal with restricted memory size, but it introduces extra computation, relies on large external Flash (\eg 32GB SD Card) and does not support general model and workload definition.
\engineshort provides complete training support for popular models at various scales including MCUNet~\cite{lin2020mcunet}, MobilenetV2~\cite{sandler2018mobilenetv2}, ResNet~\cite{he2016deep}, DistilBERT~\cite{sanh2019distilbert}, and BERT~\cite{devlin2018bert}. \engineshort optimizes both computation and memory efficiency to make on-device training easy and realistic. 

\subsection{Efficient On-Device Learning Algorithms}
Edge devices have limited computational capacity. Therefore, on-device training for edge devices often focuses on transfer learning~\cite{cai2020tinytl, kumar2022fine}. It first pre-trains the model on large-scale datasets to learn general and rich features, such as ImageNet~\cite{deng2009imagenet} for ConvNets or BooksCorpus~\cite{zhu2015aligning} for BERT. The model is then transferred to downstream tasks, such as Visual Wake Words~\cite{chowdhery2019visual} for vision or the GLUE benchmark~\cite{wang2018glue} for language. After which, the model can be customized to a small amount of personal data (\eg, learning a user's accent) to perform better at the \emph{same} task. 

Due to the smaller scale and diversity of the downstream data, people found that it is not always necessary to update the entire model to achieve a good performance. 
\emph{Sparsely} updating part of the model proves to be a good solution that achieves similar or better performance at a smaller training cost~\cite{cai2020tinytl, mudrakarta2018k, frankle2020training, lin2022ondevice, houlsby2019parameter, hu2021lora, li2021prefix}.
The most straightforward method is to fine-tune only the classifier layer~\cite{chatfield2014return,donahue2014decaf,gan2015devnet,sharif2014cnn}, but the capacity is limited when the domain shift is large. For CNN models, people have investigated fine-tuning only biases~\cite{cai2020tinytl, zaken21bitfit}, batch normalization layers~\cite{mudrakarta2019k,frankle2020training}, added parallel branches~\cite{cai2020tinytl}, \etc. The sparse backpropagation scheme is even more popular for adapting pre-trained language models (\eg, BERT~\cite{devlin2018bert}, GPT~\cite{radford2018improving}) to various downstream tasks, which significantly reduce the trainable parameters~\cite{houlsby2019parameter, hu2021lora, li2021prefix}. However, sparse backpropagation lacks system support.
Despite the great theoretical savings, existing training frameworks cannot realize measured speedup or memory saving from sparse backpropagation. \engineshort provides system-level support for such flexible workloads to deliver a faster program and efficient runtime. %

\subsection{Computation Graph Transformation and Optimizations}
There are plenty of graph transformations for inference scenarios. For example, one common transform used in edge deployment is data layout conversion, as the `NCHW' preferred by GPU training is not efficient on the edge. Another common optimization technique is layer fusion. IO-intensive layers (e.g. ReLU) can usually be fused into preceding compute-intensive layers (e.g. CONV, LINEAR).
In addition, MetaFlow~\cite{jia2019optimizing} proposes functional-preserving graph transformations to optimize DNN architectures. TASO~\cite{jia2019taso} further introduces automated generation of transformation rules using formal verification. These techniques have been proven effective in inference, but few studies have explored their performance on training, even though the training graph is much more complex.
Standing on the shoulder of conventional wisdom, \engineshort is early exploration for apply these graph optimizations techniques to on-device training and discover more potential optimizations. \engineshort shows that these optimizations bring up to 1.2x speedup.

\begin{figure*}[t]
    \centering
    \includegraphics[width=0.95\linewidth]{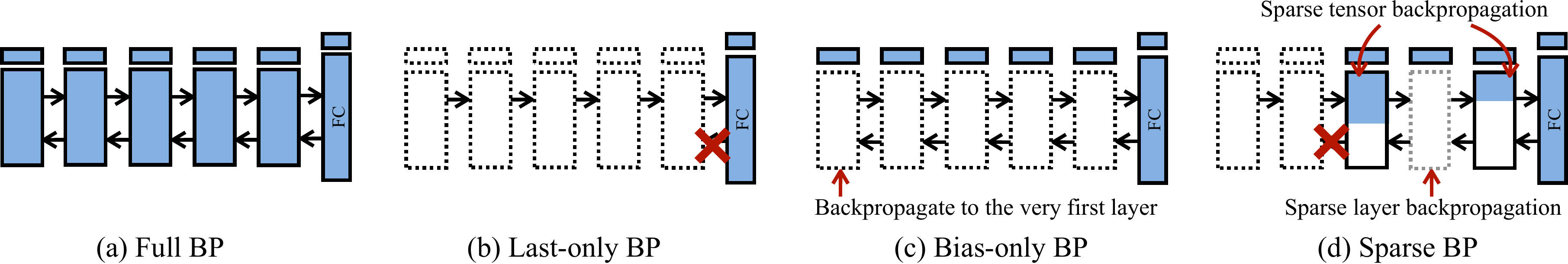}
    \caption{
    The computation graph of different backpropagation schemes on a five-layer model. We use blue to indicate the demanded intermediate activations during training. Sparse-BP delivers the best cost-quality trade-off which we will show in Section.~\ref{sec:results}.
    }
    \label{fig:update_compare}
\end{figure*}

\begin{figure*}[h]
    \centering
    \includegraphics[width=0.97\linewidth]{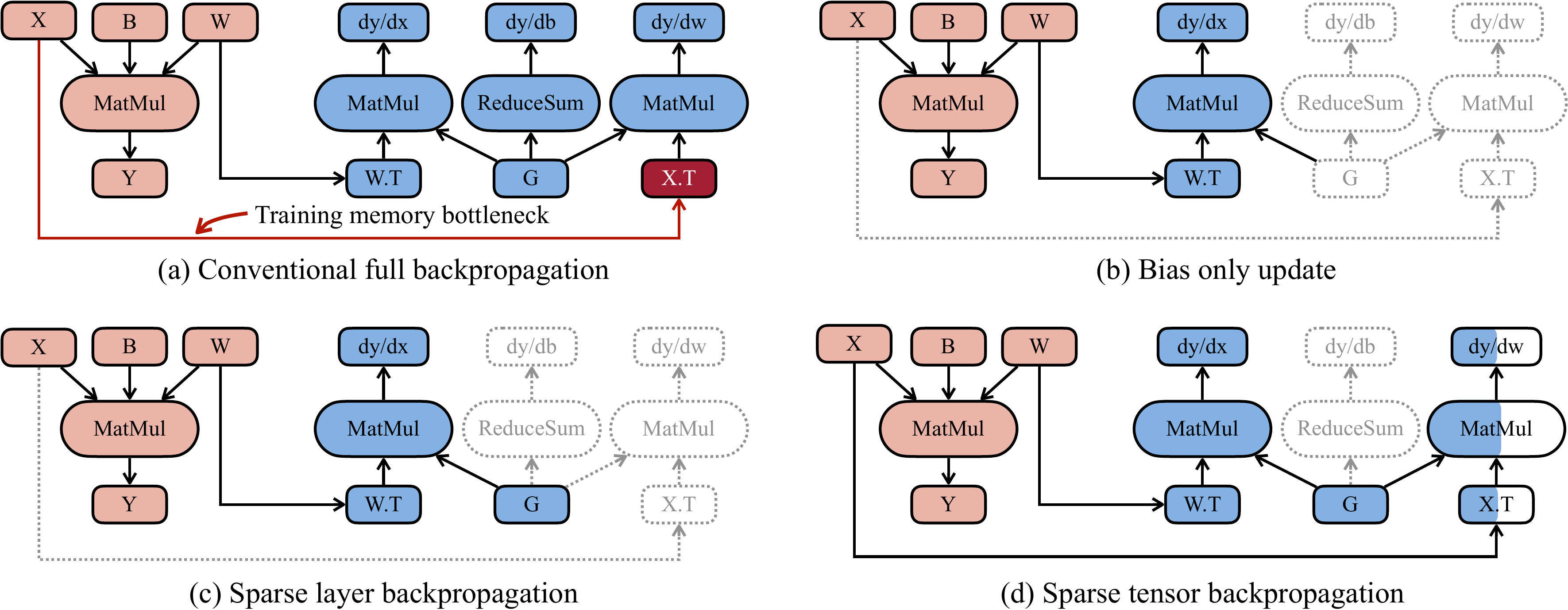}
    \caption{
    The computation graph of sparse backpropagation for a linear layer.  Red and blue blocks indicate the forward and backward OPs respectively. The red line denotes the training memory bottleneck brought by storing activations, which can be avoided using bias only / sparse update as shown in (b) (c) (d).
    }
    \label{fig:sparse_update}
\end{figure*}

\subsection{Compilation-Based Workflow}
\label{sec:method:compile-time-autodiff}
Existing training frameworks (\eg, PyTorch, TensorFlow) are based on runtime auto differentiation for flexibility. However, the design is not suitable for edge devices with limited memory and computation resources. 
Instead, \engineshort is based on a compilation-based workflow, 
sharing the following benefits:

\myparagraph{Offload Workload from Runtime to Compile Time.} 
With the compilation-centric design, we can offload part of the workload from runtime to compile time, like backward graph derivation with autodiff, memory scheduling, execution planning, \etc. Modern neural network usually consists of thousands of operators, the overhead might be small for cloud servers but not negligible for edge devices (Figure.~\ref{fig:compare_prev_ours}). 

By offloading computation to the compiler, it is possible to perform more aggressive optimizations that would not be feasible or efficient to perform at runtime. For example, \engineshort performs graph pruning, fusions, and backend switching,  which can lead to significant performance gains and memory saving.

Another advantage of compilation-based workflow is that it allows us to optimize the code across the entire program, rather than just focusing on optimizing individual operations at runtime. This not only allows us to compile used operators only to ship slim binaries, but also reveals the memory redundancy in the training loop (details in Section~\ref{section:3.3}).

\myparagraph{Support Diverse Frontends/Backends.}
Unlike the cloud, edge platforms are highly diverse, with different instruction sets, degrees of parallelism, \etc. 
Our compilation-based workflow provides general support for various frontends/backends. It can effortlessly support \emph{training} on hardware and vendor libraries that are designed specifically for \emph{inference} (\eg, \engineshort can enable training on Qualcomm Hexagon DSPs with SNPE library). 

The \emph{\engineshort frontend} takes in a neural network represented in various representations (\eg, \texttt{ONNX}, \texttt{torchscript}, \texttt{tf.graph}) and analyzes the DAG structure. It will then perform automatic differentiation (autodiff) to derive the backward graph which computes the gradients \wrt the loss function (Figure~\ref{fig:compare_prev_ours}). With the \emph{static} forward and backward graph, \engineshort will convert it into a unified intermediate representation (IR), perform graph optimizations (will be introduced later), and generate the code for different backends. Only used operators will be compiled and \engineshort link these OPs to build a light-weight executable binary.
The \emph{\engineshort backend} supports both vendor libraries (\eg, SNPE for Snapdragon GPUs and DSPs, TensorRT for NVIDIA GPUs) and customized kernels (\eg, TVM~\cite{chen2018tvm} tuning for ARM CPUs). 

Notably,  instead of binding each operator with a backward implementation (\eg, \texttt{matmul},  \texttt{matmul\_backward}), \engineshort uses the same set of primitive operations as inference to construct the training graph, allowing us to utilize inference-only backends (\eg, SNPE, TensorRT, TVM) for training, achieving high efficiency at minimal engineer effort.

\subsection{Sparse Backpropagation and Computation Graph Pruning}

Edge devices have a limited computation capacity compared to the cloud. Therefore, on-device training on edge usually targets a transfer learning/fine-tuning scenario.
Due to the smaller scale and diversity of the downstream data, people found that updating the entire model may not always lead to the best performance due to over-fitting and feature distortion~\cite{cai2020tinytl, kumar2022fine}. 
{
Updating only a subset of the models is proven to be a good solution that achieves similar or better performance at a much smaller training cost, including updating bias terms~\cite{cai2020tinytl} and the normalization layers~\cite{frankle2020training} for vision models training the low-rank parts~\cite{hu2021lora} and input prompts for language models~\cite{li2021prefix}, and sparsely update the important modules~\cite{lin2022ondevice}. 
\engineshort aims to generally support on-device training for various workloads and we focus on the sparse update to reduce training costs.
}

During the compilation, \engineshort takes in a user-defined sparse backpropagation scheme and will \emph{prune} the corresponding subgraphs of backpropagation calculation. 
\engineshort flexibly supports the following sparse backpropagation patterns:

\myparagraph{Bias-only Update.} Bias-only update does not require saving the intermediate activation~\cite{cai2020tinytl}, which significantly reduces memory usage (consider a linear layer $\mathbf{y} = \mathbf{W}\mathbf{x}$, $\mathbf{dW}=f_1(\mathbf{dy}, \mathbf{x}), \mathbf{db}=f_2(\mathbf{dy})$, only the weight gradient requires saving the input). It also saves the computation by 1/3 by skipping $\mathbf{dW}$ computation. 

\myparagraph{Layer-wise Sparse Backpropagation.} Not all the layers/weight tensors are equally important for transfer learning~\cite{lin2022ondevice}. For transfer learning to a downstream task, we find that part of the layers can be kept frozen without affecting the transfer learning performance (we can find the layers to freeze by sensitivity analysis~\cite{lin2022ondevice}; detailed in Section~\ref{sec:setups}). Therefore, we can skip the computation of part of the layers to further improve the training throughput. 

\myparagraph{Sub-layer Sparse Backpropagation.} For edge devices with limited capacity (\eg, microcontrollers), we further support sub-layer level sparse BP, where only part of the channels of a layer (convolutional layers and linear layers) are updated\footnote{Following~\cite{lin2022ondevice}, we simply update the first $k$ channels of a layer. $k$ is the \#channels to update.}. It further reduces the memory cost for storing intermediate activation (we do not need to store activation for the frozen channels) and the computation cost for gradient calculation. 
% \section{Related Work}

\section{\engine}
\begin{figure*}[t]
    \centering
    \includegraphics[width=1\linewidth]{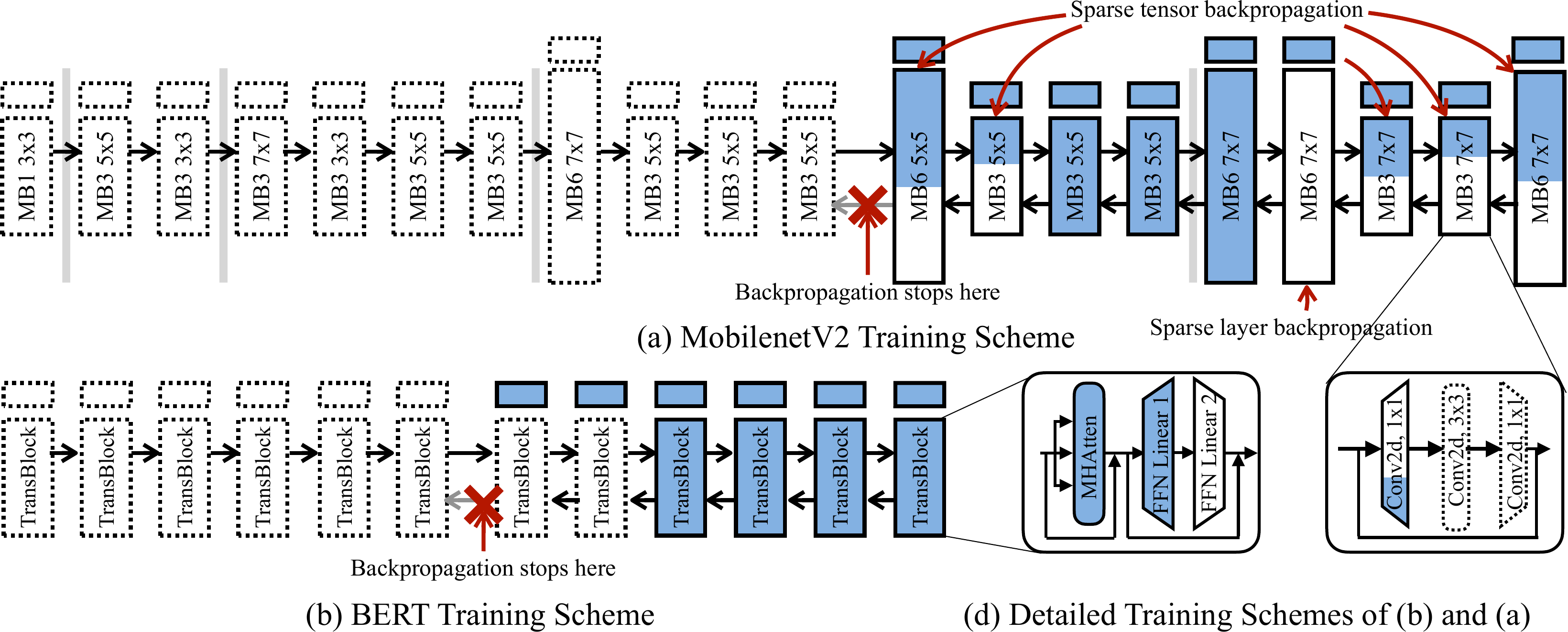}
    \caption{
    The computation graph of sparse backpropagation for ConvNet and Transformers.
    }
    \label{fig:model_sparse_update}
\end{figure*}

Compared to conventional training frameworks, sparse backpropagation has the following unique advantages
\begin{itemize}
    \item Expensive intermediate activations can be released immediately after forward  When either learning the bias-only ($dy/db$ and $dy/dx$) or fully skipping the layer (only $dy/dx$ to keep chain-rule). Thus sparse backpropagations greatly reduce the main memory bottleneck of training (the red connection line in Figure~\ref{fig:sparse_update}.a).
    \item Sparse back-propagation does not back-propagate the very first layers in DNN models since there is no need to compute gradients to the front layers if they do not require gradients (the red X mark in Figure~\ref{fig:model_sparse_update}).
\end{itemize}

None of the prior work can convert the theoretical savings into measured speed-up and memory savings. \engineshort provides systematic support for sparse BP and is able to actually reduce the on-device training cost and we expand as follows

\begin{figure}[h]
    \centering
    \includegraphics[width=0.97\linewidth]{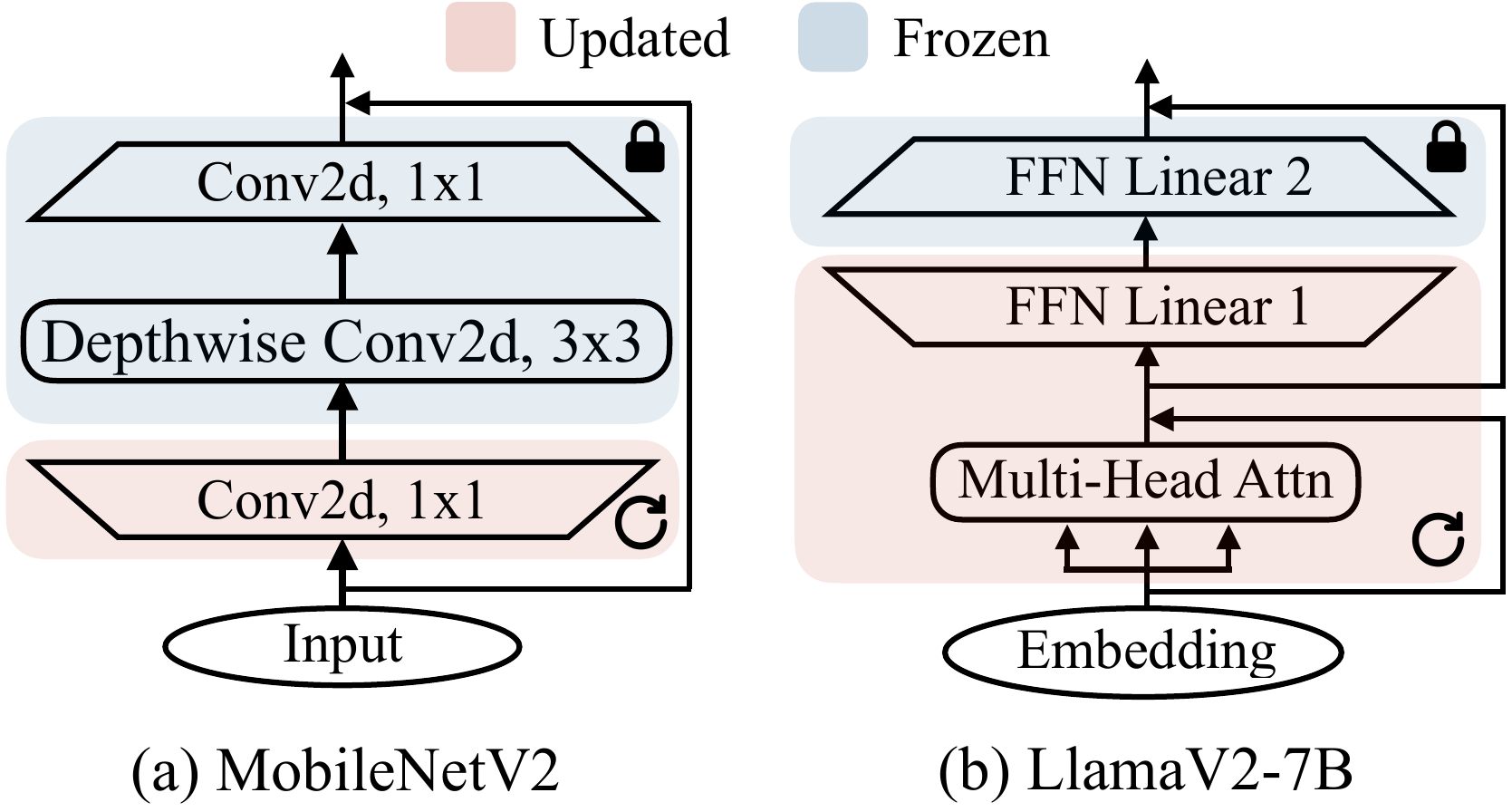}
    \caption{The sparse backpropagation schemes for MobileNetV2 and LlamaV2-7B building blocks. The first point-wise convolution plays an important role for ConvNet, while for Llama models, the attention module and first FNN layer are more important.
    }
    \label{fig:bert_mbv2_sparse_scheme}
\end{figure}

\subsection{Searching for Sparse Backpropagation Scheme}
\label{sec:method:sparsebp}

Not all the weights are equally important for transfer learning~\cite{lin2022ondevice, lee2022surgical, frankle2020training}. 
We aim to fine-tune only \textbf{the important weights} to reduce the training costs while preserving the model's accuracy. 

\myparagraph{Cost Model and Search Criterion.} 
In order to find the training scheme, we build cost models for model quality and training cost. 
Following~\cite{lin2022ondevice}, we first fine-tune only one linear (conv, fc) layer until convergence, and then repeat this process for all layers. This is an offline analysis and we use the accuracy improvement/degradation as the ``contribution'' of the weights of $i^{th}$ layer ( $\Delta\text{acc}_{\textbf{W}i}$). Similarly, we obtain the results the for bias terms of $k^{th}$ layer($\Delta\text{acc}_{\textbf{b}_k}$) and then iteratively repeat the same operations to all weights and biases to estimate their performance.

For the training cost, we focus on the memory as edge devices usually have limited memory and will easily get OOM. Thus we profile the feature map size and record it as $\text{Memory}_{k,i,r}$. We then solve the following optimization:

\begin{align}
\begin{split}
    \mathbf{k}^*, \mathbf{i}^*, \mathbf{r}^* = \max_{\mathbf{k}, \mathbf{i}, \mathbf{r}} (\sum_{\mathbf{k} \in \mathbf{i}} \Delta\text{acc}_{\textbf{b}_k} + \sum_{i\in\mathbf{i}, r\in\mathbf{r}} \Delta\text{acc}_{\textbf{W}_{i, r}}) \quad
    \\
    \text{s.t. Memory}(\mathbf{k}, \mathbf{i}, \mathbf{r}) \leq \text{constraint},
\end{split}
\end{align}
where $i$ is the layer index of weights, $k$ is the layer index of biases and $r$ is the ratio of learnable weights.  Optimizing the objectives finds the optimal update config where total contributions are maximized and the memory footprint does not exceed the constraint. We assume that the accuracy contribution of each tensor ($\Delta\text{acc}$) can be summed up thus the problem can be efficiently solved with evolutionary search. 

\myparagraph{Generalization and Acceleration} 
It is worth noting that the sparse update scheme is general and universal across different datasets. We only perform ONE scheme search on CIFAR (for vision models) and CoLA (for language models) and sparse-BP demonstrates good generalization capability. The schemes achieve competitive training accuracy compared to full fine-tuning (Table~\ref{tab:sparse_update_acc_vision} and Table~\ref{tab:sparse_update_acc_nlp}). 
Specifically, we find that for CNNs: it is most effective to update the weights of the first convolution in each block, while for transformer blocks, the weights in the attention module and the first linear layer in the Feed-Forward Network (FFN) are more important (Figure~\ref{fig:bert_mbv2_sparse_scheme}). Such schemes are also memory-efficient: the depthwise conv and second pointwise conv in the inverted bottleneck block (Figure~\ref{fig:bert_mbv2_sparse_scheme}.a) and the second linear layer in the FFN (Figure~\ref{fig:bert_mbv2_sparse_scheme}.b) have the largest input activation, while our update scheme does not require saving these large features.

After finding and specifying the gradients needed for the on-device training, \engineshort automatically traces dependency and analyzes the updated topology, then prunes the training graph using dead code elimination (DCE) to prune the computation graph and remove intermediate nodes and buffers that are no longer needed for the training.  Because the pruning is performed on graph level at compile-time, it can deliver measured memory saving and throughput improvement.

\begin{figure*}[hbt]
    \centering
    \includegraphics[width=0.97\linewidth]{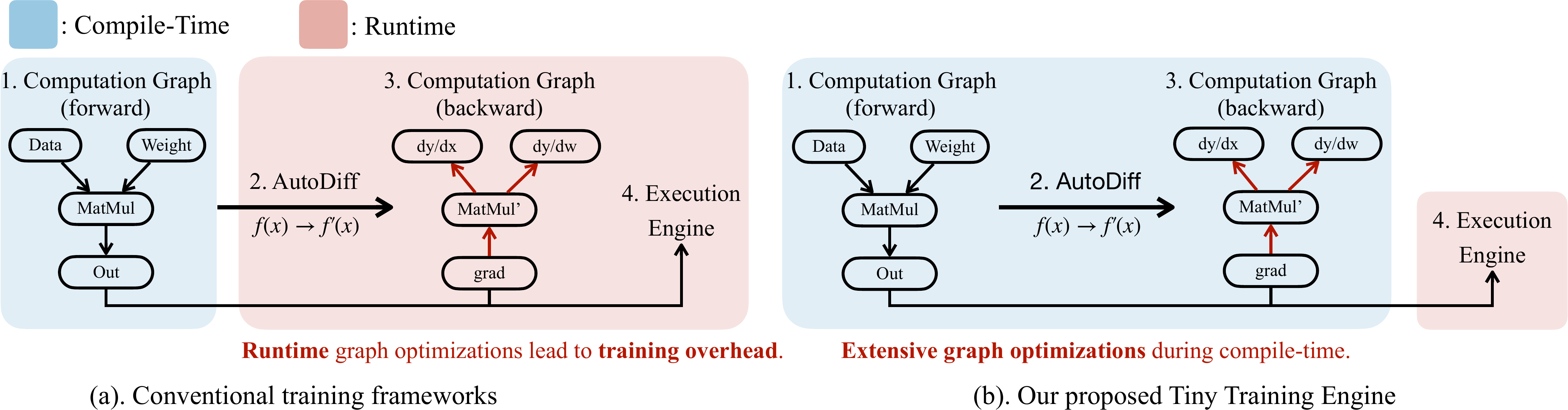}
    \caption{The comparison between runtime auto-differentiation and our compile-time differentiation. By offloading the differentiation to compile time, \engineshort not only simplifies the runtime, but also enables plenty of optimization opportunities, which will be detailed in Section.~\ref{section:3.3}. 
    }
    \label{fig:compare_prev_ours}
\end{figure*}
\subsection{Training Graph Optimization}
\label{section:3.3}
After we get the static, pruned training graph, \engineshort applies various graph optimization techniques on the unified IR before translating to different backends, which further improves the training efficiency.

\myparagraph{Operator Reordering and In-place Update.}
Different execution orders lead to different life cycles of tensors and the overall/peak memory footprint will be also affected even for the same computational graphs. This has been well-studied for inference~\cite{ahn2020ordering, liberis2019neural} but less discussed for training because the backward graph is usually derived during runtime and the compiler/scheduler does not have global information of the training process.

A concrete example is the optimizer, where the gradients are applied to update the model parameters. In conventional training, frameworks calculate all gradients and then apply the update. %
This is common among frameworks like PyTorch and TensorFlow as the \texttt{optimizer} and \texttt{forward-backward} are separate components in the system design. 
However, such a practice leads to significant memory waste for storing the gradients. %
In small batch training with sparse backpropagation, the cost of storing parameter gradients is close to peak memory usage in forward and backward as shown in Table.~\ref{tab:sparse_memory}: 
To address the overhead, \engineshort obtains all tensor information and plans for a better execution schedule. By reordering operators, the gradients can be immediately applied to the corresponding parameters before back-propagating to earlier layers. %
We further trace the life-cycle of all tensors (weights, activations, gradients) and re-order the schedules to reduce memory usage, leading up to 21x savings on microcontrollers for MCUNet. %

\textbf{Operator Fusion.}
In most deep learning frameworks, a simple operation usually requires a number of fine-grained kernels to implement. For example, a single-layer normalization operation requires three kernel calls and two memory reads and writes for forward, and six kernel calls and five memory reads and writes for backward. 
Moreover, transformations such as fusing cheap operations into expensive ones (\eg \texttt{CONV-BN-ReLU},), and parallel linear operations (\eg batch matmul) have been shown effective in improving the inference.  
During compilation and codegen, \engineshort fuse these kernels into a single one and results in less memory IO and kernel calls.

\myparagraph{Functional-Preserving Graph Transformation.}
Existing DNN frameworks optimize a computation graph by applying rules either designed by domain experts~\cite{tensorflow2015, tensorRT} or automatically discovered by program~\cite{jia2019optimizing_metaflow, jia2019taso}. There are more optimization opportunities but previous research is unable to utilize them since the backward graph was derived at runtime in earlier frameworks. Extensive investigation of potential graph optimizations will lead to slow training and incur undesired runtime overhead.

Our engine integrates these optimization techniques and is an early trial to apply to the training graph. 
\engineshort transforms the data layout for different hardware. For vision tasks, \texttt{NCHW} is the most widely used layout. But this format is only efficient on accelerators like GPU. When training on mobile CPUs / DSPs, such format is no longer optimal and \engineshort will transform the layout at compile-time to facilitate runtime training efficiency.

Furthermore, \engineshort explores different implementations of kernels. 
For example, Winograd has been widely used in inference because of its faster computation. However, the savings are not free: it requires extra pre-processing of the weights. If the weights are not static, then the transformation needs to be applied every epoch and the total FLOPs can be even higher than normal convolution. Hence it was utilized in inference and not incorporated into training frameworks. 
For on-device training scenarios, there are many frozen layers where the weights are not being changed during training~\cite{cai2020tinytl, zaken21bitfit}. These layers in fact can utilize Winograd to accelerate but such opportunities are ignored in current frameworks even if the \texttt{requires\_grad} attribute is set to \texttt{False}.
\engineshort obtains the complete training graph during compile-time thus knowing the updating information of each parameter. Therefore, we can analyze the tensor and graph information, knowing whose weights are static and whose are dynamic. \engineshort can bind operation to the fastest implementation and enable the chance to utilize Winograd even in the training. %

\begin{table*}[ht]
\centering
\caption{
    Sparse BP achieves comparable transfer learning performance $(<1\%)$ degradation on average) compared to the full update for vision models at various scales, while reducing the cost of on-device training. 
}
\small
\label{tab:sparse_update_acc_vision}
\begin{tabular}{cccccccccc}
\toprule
\multirow{2}{*}{\textbf{\begin{tabular}[c]{@{}c@{}}Vision\\ Model\end{tabular}}}       & \multirow{2}{*}{\textbf{Method}} & \multirow{2}{*}{\textbf{\begin{tabular}[c]{@{}c@{}}Avg.\\ Acc\end{tabular}}} & \multicolumn{7}{c}{\textbf{On-Device Training Vision Datasets}}                               \\ \cmidrule(lr){4-10}
                             &                         &                                                                     & Cars & CIFAR & CUB  & Flowers & Foods & Pets & VWW  \\ \midrule
\multirow{3}{*}{MCUNet-5FPS~\cite{lin2020mcunet}} & Full BP  & $74.1\%$ & $56.7 \pm{1.1}\%$ & $86.0 \pm{0.7}\%$   & $56.2 \pm{0.5} \%$ & $88.8 \pm{0.2} \%$ & $67.1 \pm{0.3} \%$ & $79.5 \pm{0.4} \%$ & $88.7 \pm{0.3} \%$ \\
        & Bias Only   & $72.7\%$ & $52.4 \pm 1.4 \%$ & $83.4 \pm 0.5\%$ & $55.2 \pm 0.6\%$ & $86.7 \pm 0.4 \%$ & $65.0 \pm 0.4 \%$ & $78.0 \pm 0.3\%$ & $88.1 \pm 0.3 \%$ \\
         & Sparse BP  & $74.8\%$ & $55.2 \pm 1.3 \%$ & $86.9 \pm 0.6\%$  & $57.8 \pm 0.4\%$ & $89.1 \pm 0.3\%$ & $64.4 \pm 0.3\%$ & $80.9 \pm 0.3 \%$ & $89.3 \pm 0.4\%$ \\ \midrule
\multirow{3}{*}{MobilenetV2~\cite{sandler2018mobilenetv2}} & Full BP & $89.2\%$ & $87.1 \pm 0.9\%$ & $96.0 \pm 0.5\%$ &   $76.6 \pm 0.8\%$  & $95.4 \pm 0.2 \%$  &  $83.9 \pm 0.2\%$   & $90.7 \pm 0.4\%$ &  $94.5 \pm 0.2\%$     \\
             & Bias Only & $87.3\%$ & $85.8 \pm 0.8\%$ & $94.0 \pm 0.7\%$ & $74.5 \pm 0.7\%$ & $95.1 \pm 0.5\%$ & $82.0 \pm 0.6\%$ & $87.6 \pm 0.5 \%$ & $92.4 \pm 0.3\%$ \\
             & Sparse BP  & $88.5\%$ & $86.4 \pm 1.0 \%$ & $95.0 \pm 0.9\%$ & $76.4 \pm 1.0\%$ &  $95.4 \pm 0.3\%$ & $81.5 \pm 0.5\%$ &  $90.4 \pm 0.3\%$ &  $94.2 \pm 0.3\%$   \\ \midrule
\multirow{3}{*}{ResNet-50~\cite{he2016deep}} & Full BP & $90.5\%$  & $88.2 \pm 0.5\%$ &    $96.8 \pm 0.4\%$  & $79.9 \pm 0.6\%$ & $94.2 \pm 0.3\%$   &     $85.2 \pm 0.4\%$    &   $93.6 \pm 0.2\%$    &  $95.3 \pm 0.1\%$     \\
     & Bias Only & $87.8\%$ & $84.3 \pm 0.6\%$ & $93.7 \pm 0.7\%$ & $75.0 \pm 0.3\%$ & $92.5 \pm 0.5\%$  & $83.7 \pm 0.3\%$ & $91.8 \pm 0.4 \%$ & $93.8 \pm 0.1\%$   \\
     & Sparse BP & $90.3\%$ & $86.7 \pm 0.7\%$ & $96.2 \pm 0.6\%$ &  $81.0 \pm 0.7\%$ & $95.6 \pm 0.3\%$ & $84.0 \pm 0.3\%$ & $93.4 \pm 0.5\%$ & $95.1 \pm 0.1\%$   \\
\bottomrule
\end{tabular}
\end{table*}

\begin{table*}[htb]
    \centering
    \caption{
        For language models, sparse BP maintains the fine-tuning accuracy for at a reduced training cost. Results are reported with mean and standard deviation for 3 runs.
    }
    \small
    \label{tab:sparse_update_acc_nlp}
    \begin{tabular}{cccccccccc}
    \toprule
    \multirow{2}{*}{\textbf{\begin{tabular}[c]{@{}c@{}}Language\\ Model\end{tabular}}} & \multirow{2}{*}{\textbf{Method}} & 
    \multirow{2}{*}{\textbf{Avg.}}
    & \multicolumn{7}{c}{\textbf{On-Device Training Language Datasets}}                                                                               \\ \cmidrule(lr){4-10}
                                    &                                  &                                                                              & CoLA & MNLI & MRPC-acc & QNLI & QQP-acc & RTE & SST-2 \\ \midrule
    \multirow{3}{*}{Distill-BERT~\cite{sanh2019distilbert}}   & Full BP                     & $76.9\%$                                                                      & $46.6\pm 1.2 \%$       & $81.9\pm 0.2 \%$       & $83.8 \pm 1.9 \%$           & $88.3\pm 0.1\%$       & $90.0\pm 0.2\%$          & $59.6\pm 1.9 \%$      & $90.8\pm 0.8\%$       \\
    & Bias Only & $72.8\% $ & $44.6\pm 0.9\%$ &	$73.2\pm 0.7\%$ & $78.9\pm2.3\%$ & $83.4\pm 1.4\%$ & $83.6\pm 0.5\%$ & $57.8\pm 2.5\%$	& $88.0\pm 1.3\%$ \\
                                    & Sparse BP                    & $77.0 \%$                                              & $47.9 \pm 1.5 \%$       & $81.1\pm 0.3\%$       & $84.2 \pm 1.8\%$           & $87.8\pm 0.1\%$       & $88.5\pm 0.3\%$          & $58.0 \pm 1.6\%$      & $90.6\pm 0.5\%$      \\ \midrule
    \multirow{3}{*}{BERT~\cite{devlin2018bert}}           & Full BP                     & $81.8 \%$                                                                      & $59.9 \pm 1.5\%$       & $84.0 \pm 0.1\%$       & $85.8\pm 1.9\%$           & $90.9\pm 0.2\%$       & $90.8\pm 0.3\%$          & $68.2\pm 2.0\%$      & $92.7 \pm 0.7\%$       \\
            & Bias Only & $78.1\%$ & $51.1 \pm 0.5 \%$ &	$78.6\pm 0.8\%$ &	$83.6\pm 2.6\%$ & $88.5\pm 1.0\%$ & $86.0\pm0.1\%$ & $67.9\pm 3.3 \%$ &	$90.7\pm 1.3\%$  \\ 
            & Sparse BP                    & $81.7\%$                                                                      & $58.6\pm 0.8\%$       & $84.4\pm 0.2\%$       & $86.2\pm 1.6\%$           & $90.8\pm 0.1\%$       & $90.3 \pm 0.6\%$          & $69.4 \pm 1.8\%$      & $91.8 \pm 0.4\%$       \\ 
    \bottomrule
    \end{tabular}
    \end{table*}

\section{Results}
\label{sec:results}
In this section, we comprehensively evaluate the performance of \engineshort. We first study the effectiveness of sparse backpropagation, then present the experimental results on different hardware and platforms, compared with other training frameworks. Finally, we discuss the graph optimization results.

\subsection{Setups}
\label{sec:setups}
\myparagraph{Models.}
We evaluate \engineshort on popular vision and language models. For vision tasks, we choose MCUNet~\cite{lin2020mcunet} (5FPS model), MobilenetV2~\cite{sandler2018mobilenetv2} (width multiplier 0.35 and 1.0), and ResNet-50~\cite{he2016deep}. All normalization layers (\eg BatchNorm) are fused into the linear operations (\eg Conv, Linear).  For masked language models, we choose the \texttt{base-uncased} version of BERT~\cite{devlin2018bert}  and DistilBERT~\cite{sanh2019distilbert} to benchmark the performance. 

\myparagraph{Datasets.}
For vision models, we first pre-trained them on ImageNet~\cite{deng2009imagenet} with resolution 224$\times$224 (except 128$\times$128 for MCUNet), and then fine-tuned on a set of downstream tasks to evaluate the transfer learning accuracy (including Cars~\cite{krause20133d}, CIFAR-10~\cite{krizhevsky2009learning}, CUB~\cite{cub}, Flowers~\cite{nilsback2008automated}, Foods~\cite{bossard2014food}, Pets~\cite{parkhi2012cats}, and VWW~\cite{chowdhery2019visual} conventional TinyML setting used in~\cite{cai2020tinytl, lin2022ondevice}). 
The NLP models (BERT and DistilBERT) are pre-trained on Wikipedia and BookCorpus~\cite{Zhu_2015_bookcorpus}. We evaluate their transfer learning performance on the GLUE~\cite{wang2018glue} benchmark (including CoLA, MNLI, MRPC, QNLI, QQP, RTE, SST-2). 
For the chatbot models, we use Llamav2~\cite{touvron2023llamav2} then fine-tuned with instructions from Stanford Alpaca dataset~\cite{alpaca}. We follow alpaca-eval~\cite{alpaca_eval} and MT-Bench~\cite{vicuna2023, zheng2023mtbench} to evaluate the response quality.

\begin{figure}[t]
    \centering
    \includegraphics[width=1.0\linewidth]{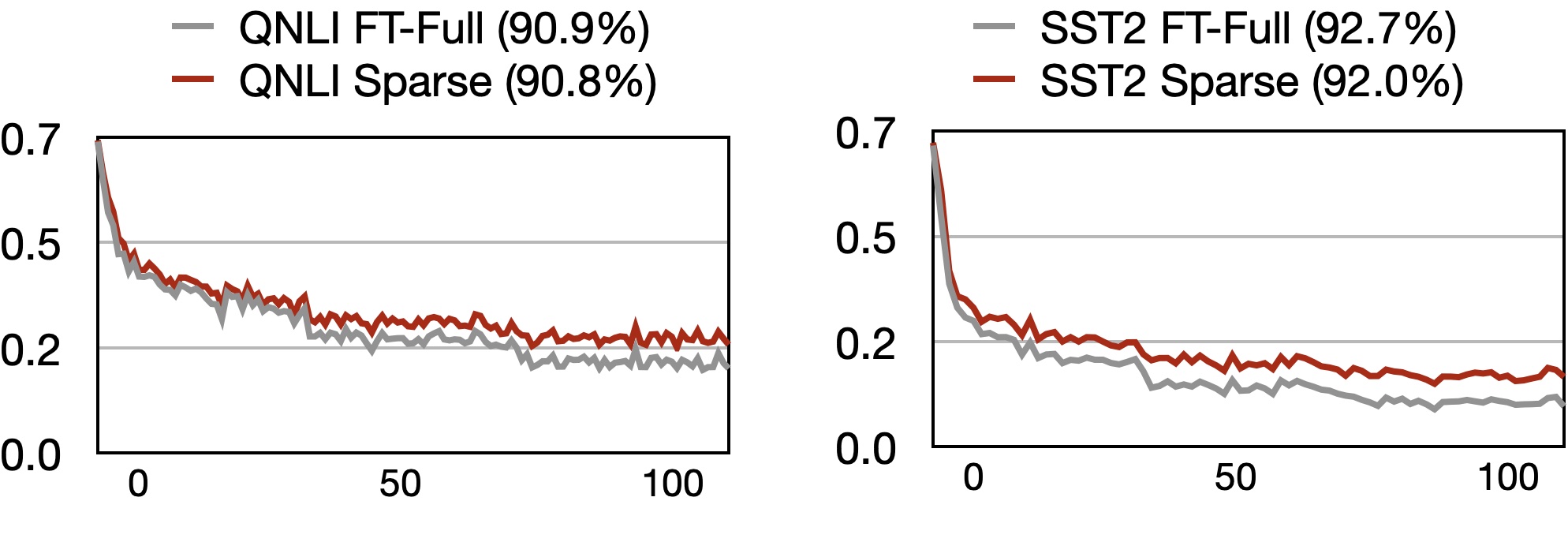}
    \caption{The training loss curves of FT-Full and our used sparse update on QNLI and SST-2 dataset using BERT. Sparse updates slightly slow down the training curve, but do not degrade the final accuracy
    }
    \label{fig:training_loss_curve}
\end{figure}

\myparagraph{Runtime adaptations.}
\engineshort is a compilation-based framework, and the compilation workflow helps to handle various frontends as well as adapt to different backends and runtimes. 
\engineshort takes models defined in PyTorch/TensorFlow/Jax and translates them into IR where graph optimizations can be applied. 
The optimized training graph is then fed to deep learning libariries like TVM~\cite{chen2018tvm}, SNPE~\cite{SNPEqualcomm} and TinyEngine~\cite{lin2021mcunetv2} to generate platform-specific binaries.

\begin{figure*}[htb]
    \centering
    \includegraphics[width=1.0\textwidth]{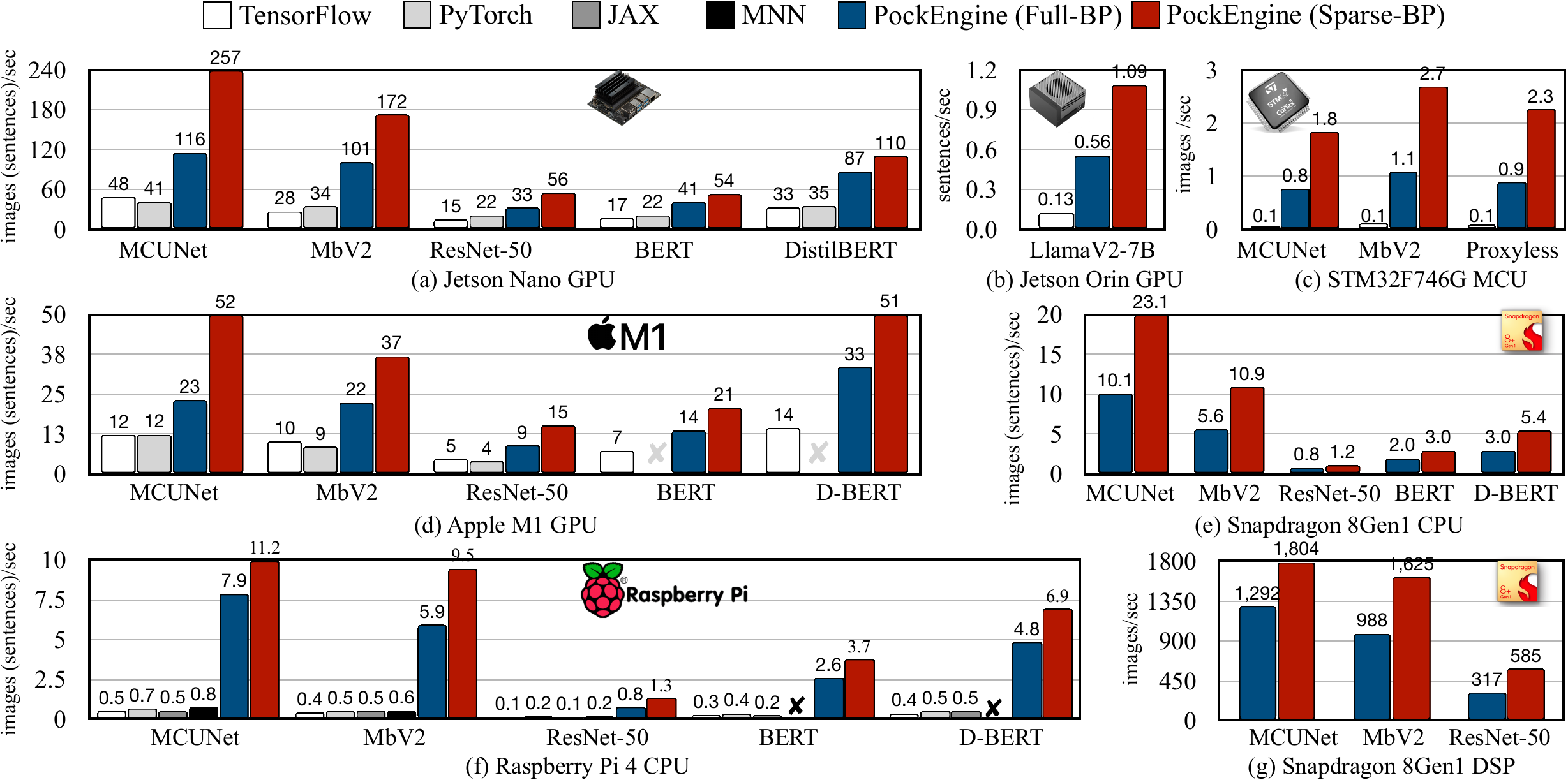}
    \caption{
        Training speed comparison between other frameworks and \engineshort of popular deep learning models on various hardware platforms. \engineshort consistently outperforms existing frameworks and sparse bp further speedups the training throughput.
    }
    \label{fig:framework_comparison}
\end{figure*}

\myparagraph{Sparse-BP Schemes for Fine-tuning.}
We obtain sparse backpropagation schemes for CNN models (MCUNet~\cite{lin2020mcunet}, MobileNet-\\V2~\cite{sandler2018mobilenetv2}, ResNet-50~\cite{he2016deep}) and transformer~\cite{vaswani2017attention}-based NLP models (BERT~\cite{devlin2018bert} and DistilBERT~\cite{sanh2019distilbert}) by sensitivity analysis~\cite{lin2022ondevice} while considering computation/memory cost. 
Our final update schemes are: 

\begin{itemize}
\item MCUNet: we further support sparse tensor backpropagation to handle the extreme memory constraints of IoT devices~\cite{lin2020mcunet}. We update the biases of the last 7 blocks and update \{100\%, 100\%, 50\%, 100\%\} of the weights of the first convolutions for the intermediate 4 blocks. 
\item MobileNetV2: update the \emph{biases} and the \emph{weights} of the first 1x1 convolution for the last 7 blocks (out of 19).
\item ResNet-50: update the \emph{biases} and the \emph{weights} of the first 1x1 convolution for the last 8 blocks (out of 16).
\item BERT: update the \emph{biases} of the last 6 blocks (out of 12) and the \emph{weights} of the attention module and the first linear in FFN for the last 4 blocks.  
\item Distill-BERT: update the \emph{biases} of the last 3 blocks (out of 6) and the \emph{weights} of the attention module and the first linear in FFN for the last 2 blocks.
\item LlamaV2-7B: update the \emph{biases} of the last 5 blocks (out of 32) and the \emph{weights} of the attention module and the first linear in FFN for the last 5 blocks.
\end{itemize}

\subsection{Effectiveness of Sparse Backpropagation}
\textbf{Sparse Backpropagation Achieves Full Accuracy.}
On-device training aims to continually improve user experience by local data thus the number of training samples is usually small compared to large-scale pre-training. Therefore, it is not necessary to perform full backpropagation to update the whole model as shown in Table~\ref{tab:sparse_update_acc_vision}. Remarkably, the downstream accuracy of sparse backpropagation can match the full-propagation baselines on both vision and language models (<1\% performance drop). On some downstream datasets, the performance of sparse backpropagation is even higher surpassing the full baselines such as Flower in vision and mrpc-acc in language. The performance is far above the common requirements for TinyML~\cite{banbury2020benchmarking} (80\% accuracy on VWW), suggesting sparse propagation is a good strategy for on-device training. 

Furthermore, when evaluating language models, sparse backpropagation also maintains the finetuning accuracy at a reduced training cost. The average performance degradation is within 1\%. This means that the use of sparse backpropagation can effectively reduce the time and cost required for training language models, without sacrificing accuracy. In fact, the results show that sparse backpropagation can even improve the model's performance on certain sub-tasks (\eg MRPC and RTE). By making training more efficient, sparse backpropagation could help to accelerate progress in these fields and enable the development of more advanced language models.

\textbf{Sparse Backpropagation Reduces Training Time and Memory.}
Besides the comparable performance when transferring to downstream tasks, sparse backpropagation greatly reduces the training peak memory and improves the training speed. 

Shown in Table~\ref{tab:sparse_memory}, the training memory grows rapidly w.r.t the batch size and soon exceeds the limitation for edge devices (\eg 1GB for Raspberry Pi), using swap or re-materialization techniques~\cite{patil2022poet} will introduce extra computation and energy cost. Sparse backpropagation cuts down peak memory usage (2.2$\times$ to 21.3$\times$) and the saving is general across models and applications. Even when batch size grows, the required memory is still small and the memory cost of training MCUNet-5FPS sparse-BP with batch size 8 is still smaller than batch 1. Batched training helps improve device utilization as well as training stability. 

When applying sparse backpropagation, operations and tensors related to frozen layers are automatically trimmed from the training graph via dead-code elimination, resulting in less computation and higher training throughput.
Figure~\ref{fig:framework_comparison} shows that the sparse backpropagation can further accelerate the speed by {1.3x to 1.6x on Raspberry Pi. Previous efficient training} algorithms only discuss the theoretical performance and \engineshort provides system-level support and translates into measured reduction.
\begin{table*}[htp]
\setlength{\tabcolsep}{12pt}
\centering
\small
\caption{
    The training memory usage comparison of full backpropagation and sparse backpropagation. We report actual memory usage measured on Jetson AGX Orin. The saving ratios are more significant as batch sizes increase. ``-'' denotes that the experiments cannot fit into devices.
}
\label{tab:sparse_memory}
\small
\begin{tabular}{ccccccc}
\toprule
\multirow{2}{*}{\textbf{Platform}} & \multirow{2}{*}{\textbf{Model}}  & \multirow{2}{*}{\textbf{\#Params}}      & \multirow{2}{*}{\textbf{Method}} & \multicolumn{3}{c}{\textbf{Training Memory }} \\   
\cmidrule(lr){5-7}
& &                                  &                                                                                  & bs=1      & bs=4      & bs=16            \\ \midrule
\multirow{2}{*}{MCU} & \multirow{2}{*}{MCUNet}                    & \multirow{2}{*}{0.6M}               & Full-BP                                                   & 3.6MB     & -     & -        \\
                        & &          & Sparse-BP                                             & 173KB      & -      & -       \\  \midrule
\multirow{2}{*}{Jetson Nano} & \multirow{2}{*}{MobilenetV2}        & \multirow{2}{*}{3.4M}      & Full-BP & 729MB & 910MB  & 1.2GB \\
   & &           & Sparse-BP  & 435MB  & 501MB &  819MB   \\      \midrule
\multirow{2}{*}{Jetson Nano} & \multirow{2}{*}{ResNet50}        & \multirow{2}{*}{26M}      & Full-BP & 827MB & 1.1GB  & 2.1GB \\
   & &           & Sparse-BP  & 663MB  & 723MB &  885MB   \\      \midrule
\multirow{2}{*}{Jetson AGX Orin} & \multirow{2}{*}{BERT}    & \multirow{2}{*}{125M}                            & Full-BP  & 1.7GB & 3.6GB & 5.7GB \\
& &            & Sparse-BP & 1.4GB & 1.9GB & 2.3GB \\ \midrule
\multirow{2}{*}{Jetson AGX Orin} & \multirow{2}{*}{LlamaV2}                    & \multirow{2}{*}{7B}               & Full-BP                                                   & 43.1GB     & -     & -        \\
& &          & Sparse-BP                                             & 31.2GB      & -      & -       \\  \bottomrule
\end{tabular}
\end{table*}

\subsection{\engineshort Speedups On-Device Training}
We compare \engineshort with other training frameworks in Figure~\ref{fig:framework_comparison}. \engineshort enables training on various hardware platforms, including Raspberry Pi 4, Snapdragon CPU and DSP, Apple M1, Jetson Nano, and microcontrollers. It also supports a wide range of models, such as MCUNet, MobilenetV2, ResNet-50, BERT, and DistilBERT. \engineshort effortlessly supports diverse models through its frontend, which converts neural networks represented in various formats to a unified intermediate representation.

Furthermore, the compilation-based workflow allows us to choose the best runtime backend for different training scenarios, including both vendor libraries (\eg SNPE for Snapdragon GPUs and DSPs, TensorRT for NVIDIA GPUs) and customized kernels (e.g., TVM-tuned kernels for ARM CPUs and Apple M1). We present a comparison of training workflows in Figure~\ref{fig:framework_comparison} and discuss it below: 

\myparagraph{Edge CPU.} For platforms like the Raspberry Pi, \engineshort offers 13 to 21 $\times$ better performance compared to popular DNN training frameworks. This speedup is due to kernel tuning, which existing frameworks either overlook in favor of GPU kernel implementations (PyTorch, TensorFlow, Jax) or optimize only for the inference pipeline and operators (MNN). The corresponding ARM kernels do not provide ideal performance, let alone the overhead brought by frameworks.

\myparagraph{Edge GPU} We benchmark edge GPU platforms using NVIDIA Jetson Nano and Jetson AGX Orin due to their widespread use in edge applications. GPUs have a much higher degree of parallelism and better training throughput than CPUs. The faster training speed of \engineshort (2.2x to 2.6$\times$ speedup) is mainly due to the compilation process: The host language Python is typically slow on low-frequency CPUs, while \engineshort's compiled graph can run without host languages. While other frameworks like TensorRT~\cite{tensorRT} may also achieve this, they are limited to inference only and do not provide training support.

\myparagraph{Apple M-Chip} The Apple M1 chip is a relatively new platform for training. While PyTorch and Tensorflow have preliminary GPU support, the compatibility is not ideal~\footnote{\url{https://github.com/pytorch/pytorch/issues/77764}}. Even with the latest build (commit ID: \textit{c9913cf}), PyTorch throws errors when launching training for BERT and DistilBERT. On the other hand, \engineshort compiles the training graph to Metal, providing better compatibility and faster training speeds.

\myparagraph{Mobile DSP}  
For Qualcomm DSP, we integrate SNPE~\cite{SNPEqualcomm} to deliver the final binaries. It is worth noting that SNPE is a conventionally inference-only library for integer models and our \engineshort easily extends it with training capability. 
As shown in Figure.~\ref{fig:framework_comparison} (g), the peak performance of DSP is impressive and even on par with edge GPUs.

\myparagraph{Microcontrollers}
For the microcontroller platform, we integrate TinyEngine~\cite{lin2020mcunet} to perform the codegen and enable training under extremely limited memory constraints. Previous frameworks like TF-Lite-Micro~\cite{tflite_micro} is \textit{inference-only} and we report the projected latency. Show in Figure.~\ref{fig:compare_prev_ours} (c), the speed is much lower than \engineshort.

\engineshort enables efficient on-device training by compilation and adaptation to various runtimes. It further supports advanced backpropagation schemes and advanced graph optimization, which we will expand further in the following section.

\begin{table*}[htp]
\centering
\small
\caption{
Instruction tuning comparisons between PyTorch and \engineshort. The pre-trained model is LLamaV2-7B~\cite{touvron2023llamav2} and we fine-tune the models following Stanford Alpaca's setting~\cite{alpaca}. We report the training loss and Alpaca-eval score~\cite{alpaca_eval} (reference model \textit{text-davinci003}. \engineshort shows significant speedup over PyTorch on Jetson AGX Orin while fully matching the training quality. With the sparse update, PockEngine further improves the training throughput while maintaining the response quality.
}

\begin{tabular}{ccccccc}
\toprule
Framework                   & Method        & \begin{tabular}[c]{@{}c@{}}Iteration\\ Latency (↓)\end{tabular}  & \begin{tabular}[c]{@{}c@{}}GPU \\ Memory(↓)\end{tabular} & Loss(↓) & \begin{tabular}[c]{@{}c@{}}Alpaca-Eval \\ Winrate(↑)\end{tabular} & \begin{tabular}[c]{@{}c@{}}MT-Bench \\ score(↑)\end{tabular} \\ \midrule
PyTorch                     & FT-Full       & 7.7s                                                             & 45.1GB                                                    & 0.761  & 44.1\%                                                            & 6.1                                                          \\
PyTorch                     & LoRA (rank=8) & 7.3s                                                            & 30.9GB                                                    & 0.801  & 43.1\%                                                            & 5.1                                                          \\ \midrule
\engineshort & FT-Full & 1.8s                                                        & 43.1GB                                                    & 0.768  & 43.7\%                                                            & 6.1                                                          \\ 
\engineshort & Sparse  & 0.9s                                                         & 31.2GB                                                    & 0.779  & 43.1\%                                                            & 5.7  \\ \bottomrule                                         
\end{tabular}
\label{tab:llms}
\end{table*}

\section{Fine-tuning ChatBot with \engineshort}
With the growing attention ChatGPT has received, the demand for fine-tuning one's own Chatbot models has also been increasing. This allows users to tailor the model to their domain-specific needs {(e.g., law, biomedical, health care) and ensures privacy (\eg private emails, personal assistant)} by not uploading information to the cloud. By fine-tuning our own language model, we can address these concerns and obtain a high-quality language model that meets our needs. In this section, we demonstrate how \engineshort can efficiently fine-tune a chatbot on edge platform {(Jetson AGX Orin)}. %

\myparagraph{Models.}
{We choose Meta's LlamaV2 ~\cite{touvron2023llamav2} and choose the 7B model} as the backbone for our experiments. This decision was based on the trade-off of model quality and device resources. The detailed fine-tuning settings are discussed below.

\myparagraph{Evaluation.}
For evaluation, we follow {Alpaca-Eval~\cite{alpaca_eval}} and MT-Bench~\cite{zheng2023mtbench} to use LLMs as the automated evaluator for benchmark generation and performance assessments.  The quality of the answers is evaluated based on helpfulness, relevance, accuracy, and details from {805 questions\footnote{\small \url{https://tatsu-lab.github.io/alpaca_eval}}} and 80 questions from  Vicuna project\footnote{\small \url{https://github.com/lm-sys/FastChat}}.
This is a pair-to-pair comparison and we choose \textit{text-davinci-003} for Alpaca-Eval win rate (\%) and \textit{ChatGPT-3.5-Turbo} for MT-Bench Score.

\myparagraph{Datasets.}
To align pre-trained language models with instructions, we follow the self-instruct~\cite{wang2022selfinstruct} and adapt data from Stanford Alpaca~\cite{alpaca}. The total training set has 52K examples containing diverse instructions and responses \footnote{\small\url{https://huggingface.co/datasets/tatsu-lab/alpaca}}

\fbox{
\small
\begin{minipage}{25em}
\textbf{Instruction:} What is the meaning of the following idiom?

\textbf{Input:} It’s raining cats and dogs.

\textbf{Output:} The idiom "it’s raining cats and dogs" means that it is raining heavily.

\begin{center}
    Example Record from Alpaca Dataset. %
\end{center}
\end{minipage}}

\myparagraph{Fine-tuning.}
We fine-tune the models for 3 epochs using a learning rate of $10^{-4}$ and no weight decay. The optimizer we use is memory-efficient Lion~\cite{chen2023symbolic}, and the maximum sentence length is limited to 512. The instruction tuning batch size is 1, and the gradient is accumulated over 16 steps. We sparsely update the biases of the last 5 blocks (out of 24) and the weights of the attention module and the first linear layer in the FFN for the last 5 blocks. We further freeze the layer-norm layers to reduce training costs and speed up training.

\subsection{Quantitative Comparison.}
\myparagraph{\engineshort Accelerates Training.}
As shown in Table ~\ref{tab:llms}, PyTorch can train on {Jetson AGX Orin}, but one iteration takes more than 7 seconds for LlamaV2-7B. Fine-tuning on 1000 records would require 2 hours while \engineshort accelerates training by 4.4x and can finish in less than half an hour.

\myparagraph{Sparse Backpropagation Accelerates Training.}
For popular parameter-efficient fine-tuning methods like LoRA~\cite{hu2021lora}, although they can effectively reduce the memory footprint (from 45.1GB to 30.9GB), the training cost is not significantly improved as they still need to backpropagate the first layer. In contrast, Sparse backpropagation reduces the backpropagation depth and significantly improves training speed (from 1768ms to 914ms, 1.9$\times$ faster).

\myparagraph{Sparse-BP Achieves Comparable Accuracy.}
Besides training throughput improvement, sparse backpropagation also maintains fine-tuning accuracy. When compared to full-BP, sparse-BP demonstrates similar performance, achieving an impressive Alpaca-Eval score of 43.7. This score closely matches the performance of full-BP, which has an Alpaca-Eval score of 44.1. Sparse-BP also performs favorably when compared to LoRA (Alpaca-Eval score 43.1).

\subsection{Qualitative Comparison between Full-BP and Sparse-BP}
In addition to quantitative analysis on accuracy and throughput, we also conduct qualitative analysis to compare the quality of sparse-BP and full-BP when performing instruction tuning on LlamaV2-7B. We make the assumption that a model's knowledge and capabilities are learned primarily during pretraining, while alignment teaches LLMs how to interact with users~\cite{zhou2023lima}. Therefore, we examine the response quality from multiple perspectives:

\myparagraph{Factual Recall}:
To thoroughly evaluate the performance between Full- and Sparse-backpropagation, we conduct a detailed analysis of the corresponding models' proficiency to answer questions related to general knowledge.
We first examine the capability to correctly identify the capital city of a given country. All models consistently deliver accurate responses while the tuned models not only provide accurate answers but also offer more comprehensive information, enriching the overall user experience.

\fbox{
\small
\begin{minipage}{25em}

\textbf{Input:} What is the capital of United Status?

\textbf{LlamaV2-7B-original:}  Washington, D.C.

\textbf{LlamaV2-7B-tuned (Full-BP):} The capital of the United States is Washington D.C. (District of Columbia).

\textbf{LlamaV2-7B-tuned (Sparse-BP):} The capital of the United States is Washington, D.C.
\begin{center}
    Example 1.The Capital of U.S.A.
\end{center}
\end{minipage}
}

Similarly, when we inquire about the dates of notable events, all models exhibit a remarkable level of confidence and accuracy in their responses. This suggests the reliability and correctness of both Full- and Sparse-backpropagation during fine-tuning.

\fbox{
\small
\begin{minipage}{25em}

\textbf{Input:} In what year was the Beijing Olympics held?

\textbf{LlamaV2-7B-original:}  The Beijing Olympics were held in 2008..

\textbf{LlamaV2-7B-tuned (Full-BP):} The Beijing Olympics, also known as the Games of the XXIX Olympiad, were held in 2008. The opening ceremony took place on August 8, 2008, and the closing ceremony was held on August 24, 2008.

\textbf{LlamaV2-7B-tuned (Sparse-BP):} The Beijing Olympics, also known as the 2008 Summer Olympics, were held in Beijing, China in 2008.

\begin{center}
    Example 2. The Date of Beijing Olympics.
\end{center}
\end{minipage}
}

However, when the question becomes obscure, the original Llama model no longer provides accurate answers. For example, the raw Llama model (without fine-tuning) incorrectly states that Michael Jackson died in 2020 (the actual year is 2009), and it also provides incorrect album names (the proper title is ``Invincible''). Meanwhile, the Full-BP model and the Sparse-BP model both properly interpret the questions and provide accurate responses.

\fbox{
\small
\begin{minipage}{25em}
\textbf{Input:} what was Michael Jackson's last album?

\textbf{LlamaV2-7B-original:}  Michael Jackson passed away on June 25, 2020, so he did not release an album after that. However, before his death, he released the album "Michael Jackson: The Ultimate Collection," which was a compilation of his previous studio albums and bonus tracks.

\textbf{LlamaV2-7B-tuned (Full-BP):} 
Michael Jackson's last album was 'Invincible', released in 2001.

\textbf{LlamaV2-7B-tuned (Sparse-BP):} 
Michael Jackson's last album, released during his lifetime, was "Invincible" which was released in 2001. 

\begin{center}
    Example 3. The Title of Michael Jackson's Final Album
\end{center}

\end{minipage}}

\myparagraph{Operation and Logic Inference}
Next, we evaluate a simple but interesting task: reverse the order of words. This task requires LLMs to understand the concept of reversing and properly segmenting the sentences.
The original model misunderstood the task and reversed the words' meanings, leading to incorrect answers. Meanwhile, the Full-BP and sparse-BP models, which were fine-tuned, both correctly understood the term "reverse" and executed the given task, though the sparse-BP model made a tiny mistake and still misplaced the word "the".

\fbox{
\small
\begin{minipage}{25em}
\textbf{Input:} Please reverse the words in the sentence "I love the Micro conference"

\textbf{LlamaV2-7B-original:}  I hate the conference Micro.

\textbf{LlamaV2-7B-tuned (Full-BP):} 
The reversed result is "Conference Micro the love I"

\textbf{LlamaV2-7B-tuned (Sparse-BP):}
The reversed sentence would be "Conference the Micro love I".

\begin{center}
    Example 4. Reversing a sentence.
\end{center}
\end{minipage}}

After comparing the response quality of executing given tasks, we then evaluated the logical inference capability of the fine-tuned models. We first briefly introduced the hosting location of the Micro conference for the past three years and asked where the conference was held last year. Without fine-tuning, the original Llama model failed to understand the question and only simply repeated the first sentence. However, for the fine-tuned models, both Full-BP and Sparse-BP were able to correctly name the place, with the response from Full-BP being more confident.

\fbox{
\small
\begin{minipage}{25em}
\textbf{Input:} Micro was running virtually online in 2021, and held physically at Chicago in 2022 and will soon be hosted in Toronto in 2023. Now is the year of 2023, where was Micro hosted last year?

\textbf{LlamaV2-7B-original:}  
Micro was hosted virtually in 2021

\textbf{LlamaV2-7B-tuned (Full-BP):} 
Micro was hosted physically in Chicago in 2022.

\textbf{LlamaV2-7B-tuned (Sparse-BP):}
Based on the information provided, it seems that Micro was hosted physically at Chicago in 2022.

\begin{center}
    Example 5. Inferring Information Based on Context.
\end{center}
\end{minipage}}

\myparagraph{Maths}

We then compared the mathematical abilities of different models. Mathematics is often considered a significant challenge for many language models, and obviously, the original LlamaV2 model struggled with math-related questions, while the Full-BP and Sparse-BP models demonstrated certain capabilities for reasoning and properly calculating the results.

\fbox{
\small
\begin{minipage}{25em}
\textbf{Input:} Lee mows one lawn and charges \$33. Last week he mowed 16 lawns and three customers each gave him a \$10 tip. How many dollars did Lee earn mowing lawns last week?

\textbf{LlamaV2-7B-original:}  Lee earned \$660 from mowing lawns last week.

\textbf{LlamaV2-7B-tuned (Full-BP):} 
Lee earned \$33 x 16 = \$528 from mowing lawns. His total earnings from the 16 customers who gave him a \$10 tip is \$10 x 3 = \$30. Therefore, Lee earned a total of \$528 + \$30 = \$558 last week mowing lawns.

\textbf{LlamaV2-7B-tuned (Sparse-BP):} 
To determine how much Lee earned mowing lawns last week, we need to calculate the total amount of money he earned from mowing lawns and tips.
Lee mowed one lawn and charged \$33, so he earned \$33 from that lawn.
Last week, Lee mowed 16 lawns, and three customers each gave him a \$10 tip, so he earned \$10 x 3 = \$30 in tips from those lawns and earned \$33 from mowing one lawn and total 16 lawns = \$528.
Therefore, Lee earned \$558 last week from mowing lawns and receiving tips.

\begin{center}
    Example 6. Math Problem Solving
\end{center}
\end{minipage}}

Note that this is a concise qualitative study comparing original, Full-BP fine-tuned, and Sparse-BP fine-tuned LLMs. We carefully selected representative samples for this study, although it is important to note that it is not comprehensive given the extensive range of responses the model can provide. The objective of this analysis is to present compelling evidence in support of two findings: (1) fine-tuning is an essential process for personalizing your own Chabot, and (2) Sparse-BP is capable of fine-tuning models with comparable quality with much reduced cost.

\section{Conclusion}
\label{sec:conclusion}

We present \engine, an efficient training framework for learning on edge. \engineshort has general support for various frontends/backends to deal with hardware heterogeneity on edge. It improves the efficiency of on-device training via (1) compilation-based auto-differentiation to offload overheads from runtime to compile time; (2) supporting sparse backpropagation with backward graph pruning; (3) training graph optimization including operator reordering/fusion and various function-preserving transforms. 

Experiments on different edge devices show \engineshort can significantly speedup on-device training: 11.2$\times$ on ARM CPUs, 2$\times$ on Apple M1, and 2.7$\times$ on NVIDIA edge GPU, and 9.6$\times$ on microcontroller compared to TensorFlow. \engineshort supports sparse backpropagation, which further speeds up by 1.5 - 3.5 $\times$ while matching the accuracy of full backpropagation. 
Further, PockEngine enables fine-tuning {LLamaV2-7B language model} on a Jetson AGX Orin at 914ms, 7.9$\times$ faster than the PyTorch baseline.
We hope our engine design can facilitate AI applications with personalization and life-long learning capacity by democratizing learning on the edge.

\begin{acks} 
This work was supported by MIT-IBM Watson AI Lab, MIT AI Hardware Program, MIT-Amazon Science Hub, and NSF. Ligeng Zhu and Ji Lin were partially supported by the Qualcomm Innovation Fellowship.
\end{acks}

\balance
\bibliographystyle{ACM-Reference-Format}
\bibliography{main}


\end{document}